# Neural Network Driven, Interactive Design for Nonlinear Optical Molecules Based on Group Contribution Method

Jinming Fan, Chao Qian, Shaodong Zhou*

*Mr. Jinming Fan, Prof. Dr. Chao Qian, and Dr. Shaodong Zhou*

*College of Chemical and Biological Engineering, Zhejiang Provincial Key Laboratory of Advanced Chemical Engineering Manufacture Technology, Zhejiang University, 310027 Hangzhou (P. R. China)*

*Zhejiang Provincial Innovation Center of Advanced Chemicals Technology, Institute of Zhejiang University - Quzhou, 324000 Quzhou (P.R. China)*

E-mail: szhou@zju.edu.cn

***ABSTRACT:*** *A Lewis-mode group contribution method (LGC) – multi-stage Bayesian neural network (msBNN) – evolutionary algorithm (EA) framework is reported for rational design of D-π-A type organic small-molecule nonlinear optical materials is presented. Upon combination of msBNN and corrected Lewis-mode group contribution method (cLGC), different optical properties of molecules are afforded accurately and efficiently – by using only a small data set for training. Moreover, by employing the EA model designed specifically for LGC, structural search is well achievable. The logical origins of the well performance of the framework are discussed in detail. Considering that such a theory guided, machine learning framework combines chemical principles and data-driven tools, most likely, it will be proven efficient to solve molecular design related problems in wider fields.*

## INTRODUCTION

Data driven machine learning[1-2] continues to play important roles in the field of materials[3-5] and chemistry.[6-9] A training model with sufficient samples can be used to train for accurate prediction of various molecular properties, which also dramatically accelerates the discovery of new functional molecules.[10-13] For example, unsupervised generative model based on Variational autoencoder (VAE)[14] and adversarial neural networks (GAN)[15] can quickly generate a large number of molecules; upon combination with prediction models, the reverse molecular design is achievable. However, this model requires a large amount of data for training in order to achieve satisfactory results. When expanding into new fields, insufficient experimental data and a lack of reasonable methods limit the feasibility of the above model, and people may only rely on quantum mechanical calculations.[16] In the field of organic nonlinear optical (NLO) materials, small molecules with D-π-A structure as the framework have made certain progress based on experiments, but the discovery of such molecules traditionally relies on the experience of chemists to design new molecules by combining various functional groups.[17-18] However, such an experimental synthesis - DFT calculation validation strategy suffers from low efficiency.[19] Therefore, to assist chemists in conducting experiments more efficiently, an urgent requirement for faster molecular design with limited data thus emerges.

Group contribution (GC) is a classic method traditionally used to predict thermodynamic properties.[20-22] However, the cost and difficulty of manual modeling limit its development in optical properties. Driven by big data, deep learning converts the representation of sample features in the original space into a new representation through feature conversion stage by stage, which makes classification or prediction easier.[23] The problem is that, without theoretical guidance a high accuracy relies only on the accumulation of data. Interestingly, an elaborate combination of the GC method and machine learning (ML) model is likely to touch the essence of spectroscopic features.[24] Accordingly, the applicability of the GC method is thus expanded, while the interpretability of ML models enhanced. Upon further investigation, we found that the GC-ML strategy may serve as the very ideal tool for efficient design of organic NLO molecules.

Herein, we report a Lewis-mode GC method (LGC) and its corrected version (cLGC) based on the characteristics of small organic molecules of D-π-A structure. Based on this method, a target-driven molecular design framework combining Bayesian neural network and evolutionary algorithm (EA)[25-26] is designed. The framework starts from stage-by-stage evaluation of group contribution to optical properties thus eventually affording integrated molecular feature, and subsequently embeds them into the evolutionary algorithm. The latter has been adjusted specially for LGC methods and is able to mutate independently to cross different molecular regions; accordingly, the target structure of desired can be obtained. Afterwards, the cLGC- msBNN model will make more accurate predictions of the optical properties of the obtained molecules (Figure 1d). This framework concerts chemical principles and algorithmic concepts to achieve accurate prediction and molecular screening within a regulated group space based on only a small amount of data.

# RESULTS AND DISCUSSIONS

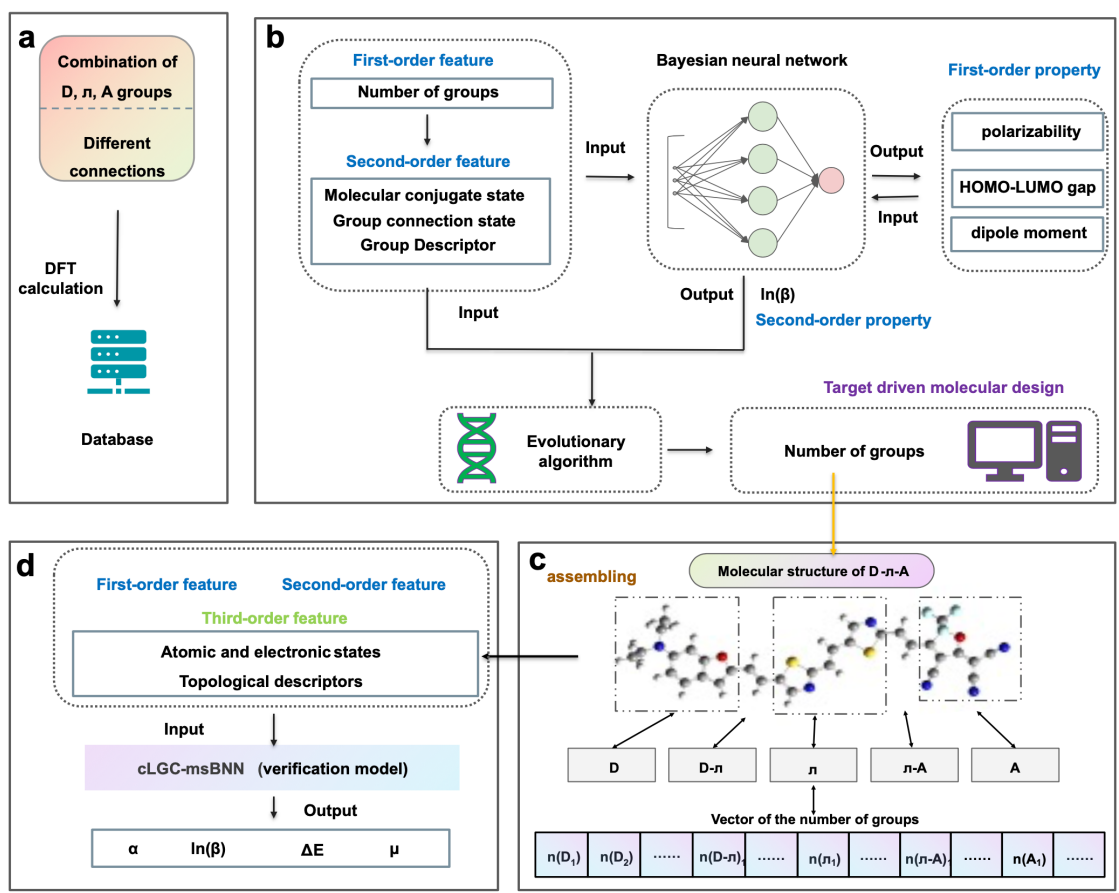

**Figure 1.** Flow chart of the framework as developed in this work. (a) Design and calculation of data. (b) Working principle diagram of LGC-msBNN model and evolutionary algorithm.(c) The working principle diagram of the group contribution method. (d) Working principle diagram of cLGC-msBNN model.

**Data preparation and Feature engineering**

Data feature processing and data selection play a crucial role in prediction and model application. Therefore, it is particularly important to develop reasonable feature inputs based on the target property. For D-π-A type NLO materials, there are several factors that affect their photoelectric properties: 1) Different combinations of electron donor group, electron acceptor group and the bridge group;[26] 2) The bond length alternation (BLA);[27-29] 3) The degree of molecular conjugation.[30] However, when simple aromatic ring participates in the conjugation of electron bridges, BLA becomes ambiguous, so a method is needed to quantitatively describe the degree of such alternation. To this end, we have developed a Lewis-mode group contribution method (LGC), which is based on the D-π-A backbone. As shown in Figure 1b, this method clearly classifies the composition of molecular structures and the connection mode between different aromatic groups, thus enabling accurate prediction of optical properties by counting contribution from each constituent. Moreover, another problematic aspect, i.e. the nexus between different groups, is also soluble via classification of the groups regarding their donor/acceptor nature. Accordingly, the reverse molecular design process is facilitated. However, the LGC method considers only the structural features at the molecular level. In order to further describe the contribution of electronic features like electrostatic interaction between atoms, valence states of the elements, the bonding state of each atom, and other factors (the third-order features) influential to optical properties, we further corrected the Lewis-mode GC method (cLGC) (for more details, see DATA AND METHOD section).

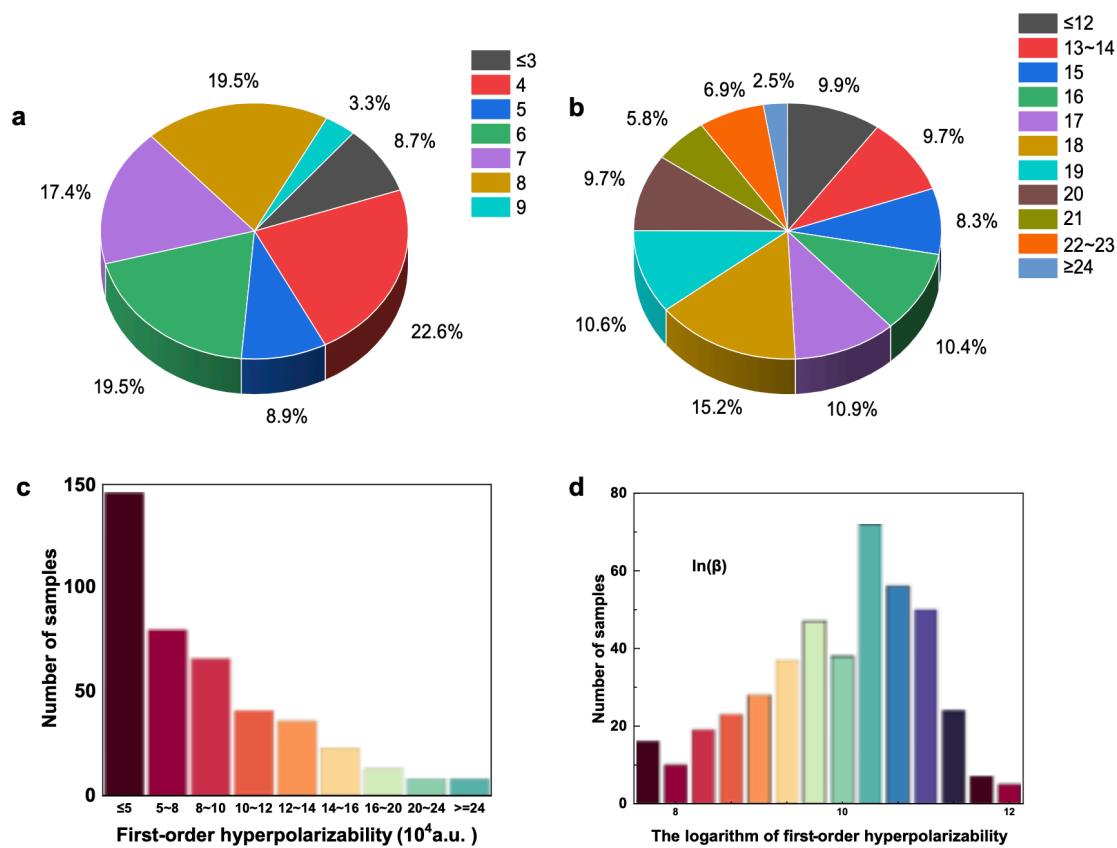

**Figure 2**. Sample features of the dataset. (a) Distribution of conjugated double bond numbers of electron bridge groups. (b) The Distribution of conjugated double bond numbers in D-π-A type Molecules. (c) Distribution of first-order hyperpolarizability (β). (d) Distribution of the logarithm of first-order hyperpolarizability (ln(β)).

Based on the LGC method, the required electronic features for training were obtained using DFT calculation of 429 molecules. Partial information of these molecules is shown in Figure 2 and Figure 3. We considered not only the combination of different electron donor and acceptor groups, but also the combination of simple aromatic ring and single/double bonds in the electronic bridge (Figure 3). Thus, our dataset includes electron donors/acceptors and bridges of different conjugation lengths (Figure 2a and 2b).

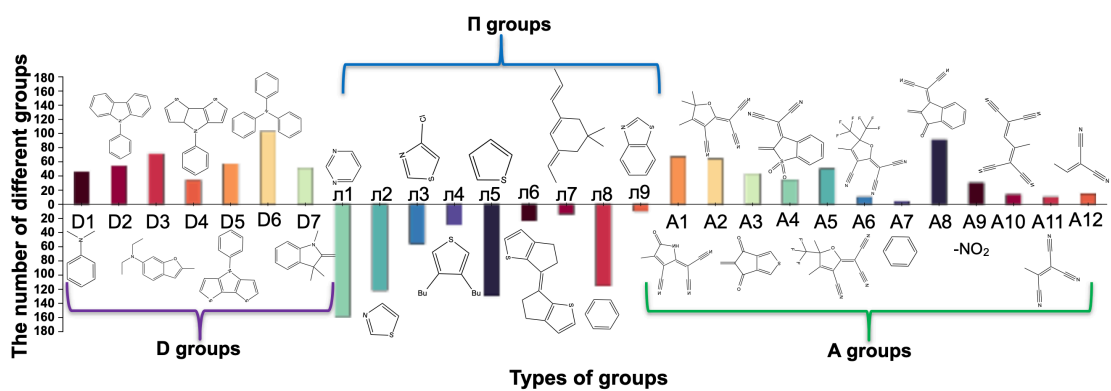

**Figure 3**. The number of electron donor groups (D groups), electron bridge groups (л groups), and electron acceptor groups (A groups) used in different categories in the dataset.

As a key property of D-л-A type molecules, the first-order hyperpolarizability (β) is distributed mainly at low values (Figure 2c). To improve the performance of the prediction model, the logarithm of first-order hyperpolarizability ln(β) was used as the target feature (Figure 2d).

**Model development**

For large datasets, deep learning makes it easy to identify connections from a wide range of features and output results. However, for small sample learning, one-pot stew for modeling is unreliable. Therefore, one needs to indicate the direction for ML models with chemical theory, which is the LGC method designed for. As shown in Figure 1b, different groups of D-л-A type molecules will be represented by independent group vectors (the first-order characteristics): $\vec{n_\pi}, \vec{n_{D-\pi}}, \vec{n_{\pi-A}}, \vec{n_A}, \vec{n_D}$.

Next, the degree of conjugation and the overall structure greatly affect the first-order hyperpolarizability. For the LGC method, each D, A and л group is grant with the same degree of conjugation, the degree of group conjugation ($C_\pi, C_D, C_A$) and molecular conjugation ($C$) can be calculated based on the number of groups (Equations (1) – (4)):

$$C_\pi = \vec{w_\pi} * \vec{n_\pi} + \vec{w_{D-\pi}} * \vec{n_{D-\pi}} + \vec{w_{\pi-A}} * \vec{n_{\pi-A}} \quad (1)$$

$$C_D = \vec{w_D} * \vec{n_D} \quad (2)$$

$$C_A = \vec{w_A} * \vec{n_A} \tag{3}$$

$$C = C_\pi + C_D + C_A \tag{4}$$

in which $\vec{n}$ is vector of the number of groups with different structures, $\vec{w}$ is the conjugate weight vectors of each group. Next, the contribution of secondary groups (e.g. the number of -C≡N groups in electron acceptor groups, the number of benzene rings, etc.) is considered in combination with the descriptors of molecules (e.g. connection mode of aromatic ring) (Equations (5) – (8)):

$$\vec{n_{A\_des}} = \vec{n_A} * \vec{w_{A_{des1}}}, \vec{n_A} * \vec{w_{A_{des2}}}, \cdots \tag{5}$$

$$\vec{n_{A\_g}} = \vec{n_A} * \vec{w_{A_{g1}}}, \vec{n_A} * \vec{w_{A_{g2}}}, \cdots \tag{6}$$

$$\vec{n_{des}} = (\vec{n_{A\_des}}, \vec{n_{\pi\_des}}, \vec{n_{D\_des}}) \tag{7}$$

$$\vec{n_g} = (\vec{n_{A\_g}}, \vec{n_{\pi\_g}}, \vec{n_{D\_g}}) \tag{8}$$

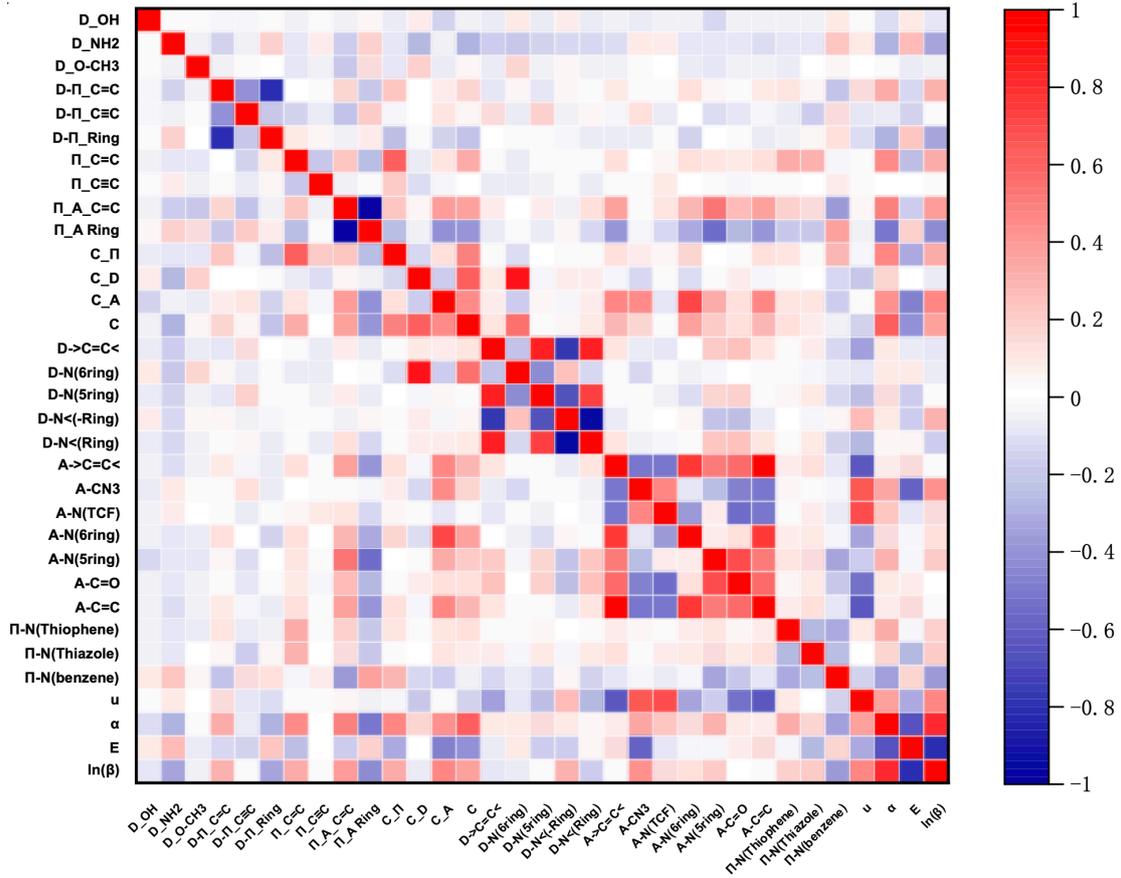

**Figure 4**. The heat map of the Pearson correlation coefficient matrix between the features.

in which $\overrightarrow{w_{A_{des}}}$, $\overrightarrow{w_{A_g}}$ are the weight vectors of group descriptors and second-order groups, respectively, for each electron acceptor group (A group); $\overrightarrow{n_{des}}$, $\overrightarrow{n_g}$ are the vectors of the group descriptors and second-order groups for the entire molecule. As shown in Figure 4, there is a direct correlation between second-order features and optical properties, which will be beneficial for their prediction. Moreover, as shown in Figure 5a, there is a strong correlation between first-order optical properties and ln (β), after completing the feature processing of the entire molecule, four *Bayesian* neural network models (BNN) were used to predict the optical properties stage by stage (Equations (9) – (14)):

$$\overrightarrow{F_1} = (\overrightarrow{n_\Pi},\ \overrightarrow{n_{D-\Pi}},\ \overrightarrow{n_{\Pi-A}},\ \overrightarrow{n_A},\ \overrightarrow{n_D}) \quad (9)$$

$$\overrightarrow{F_2} = (C_\Pi,\ C_D,\ C_A,\ C,\ \overrightarrow{n_{des}},\ \overrightarrow{n_g}) \quad (10)$$

$$\mu = \text{BNN1}(\vec{F_1}, \vec{F_2}) \tag{11}$$

$$\alpha = \text{BNN2}(\vec{F_1}, \vec{F_2}) \tag{12}$$

$$\Delta E = \text{BNN3}(\vec{F_1}, \vec{F_2}) \tag{13}$$

$$\ln(\beta) = \text{BNN4}(\vec{F_1}, \vec{F_2}, \alpha, \Delta E, \mu) \tag{14}$$

in which $\mu$, $\alpha$, and $\Delta E$ represents dipole moment, polarizability, and the HOMO-LUMO gap, respectively. $\vec{F_1}$ and $\vec{F_2}$ represents first-order molecular features and second-order molecular features. Thus, a series of second-order molecular features will be calculated for D-π-A type molecules, and their first-order optical properties will be predicted through three neural network models, ultimately affording first-order hyperpolarizability.

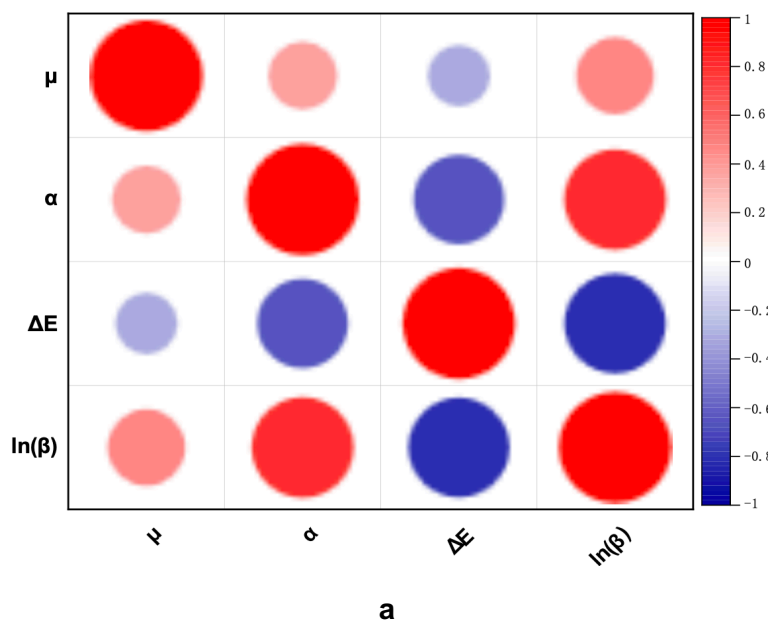

a

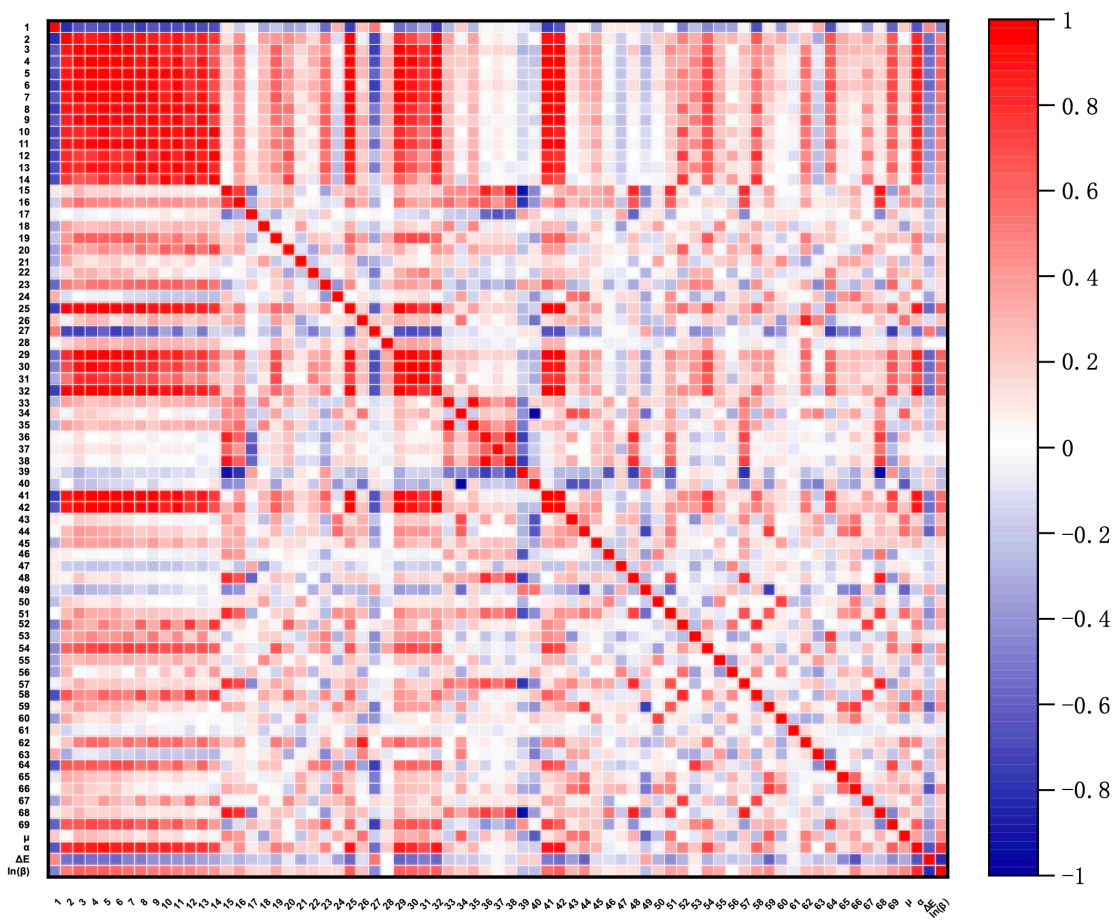

**b**

**Figure 5.** The heat map of the Pearson correlation coefficient matrix between the features. (a) The correlation between first-order optical properties and second-order optical properties. (b) The correlation between third-order features and optical properties.

The cLGC-msBNN model and LGC-msBNN model share the same working principle, with the only difference being the input features. (Equations (15)):

$$\ln(\beta) = \text{msBNN}(\vec{F_1}, \vec{F_2}, \vec{F_3}) \qquad (15)$$

in which msBNN represents a multi-stage neural network composed of BNN1, BNN2, BNN3, and BNN4. $\vec{F_3}$ represents the newly added third-order feature of cLGC, as shown in Figure 5b, most of the third-order features we add directly affect the optical properties of molecules.

To reflect the real performance of the model, instead of "carefully" selected result, the average of 100 randomly train-test-split results were used as the evaluation indicator of the model. Compared to ten-fold cross validation, this method is more randomly, with each data being used multiple times as a test set for prediction.

Table 1. Prediction results of different molecular properties using the LGC-BNN model and cLGC-BNN model.

| Method | LGC-BNN | | | | cLGC-BNN | | | |
|---|---|---|---|---|---|---|---|---|
| property | Test set | | | | | | | |
| | MAE | MRE | R | R2 | MAE | MRE | R | R2 |
| α (Bohr**3) | 30.9055 | 0.0480 | 0.9597 | 0.9215 | 24.0565 | 0.0370 | 0.9775 | 0.9556 |
| ΔE (ev) | 0.1064 | 0.0601 | 0.9201 | 0.8482 | 0.0954 | 0.0549 | 0.9460 | 0.8954 |
| μ (D) | 1.3327 | 0.1730 | 0.9337 | 0.8724 | 1.3034 | 0.1675 | 0.9328 | 0.8707 |
| | Train set | | | | | | | |
| | MAE | MRE | R | R2 | MAE | MRE | R | R2 |
| α (Bohr**3) | 23.7264 | 0.0365 | 0.9780 | 0.9565 | 19.9927 | 0.0309 | 0.9854 | 0.9710 |
| ΔE (ev) | 0.0797 | 0.0490 | 0.9614 | 0.9243 | 0.0764 | 0.0433 | 0.9669 | 0.9349 |
| μ(D) | 1.0503 | 0.1376 | 0.9606 | 0.9228 | 0.9500 | 0.1252 | 0.9665 | 0.9341 |

Table 2. Prediction results of ln(β) using different methods.

| Method | Test set | | | |
|---|---|---|---|---|
| | MAE | MRE | R | $R^2$ |
| LGC-msBNN | 0.2073 | 0.0192 | 0.9256 | 0.8560 |
| cLGC-msBNN | 0.1931 | 0.0179 | 0.9390 | 0.8825 |
| | Train set | | | |
| | MAE | MRE | R | $R^2$ |
| LGC-msBNN | 0.0793 | 0.0072 | 0.9901 | 0.9803 |
| cLGC-msBNN | 0.0738 | 0.0069 | 0.9924 | 0.9848 |

Firstly, we compared the predictive performance of LGC-BNN and cLGC-BNN for the first-order optical properties, i.e. the mean absolute error (MAE), average percentage error (MRE), Pearson correlation coefficient(R), and the square of Pearson correlation coefficient (R2) of the prediction results (Table 1). It turned out that both methods afford high accuracy without significant overfitting, which ensures that the performance of the models will not be seriously affected due to the accumulation of errors in practical application (Table 2).

Figure 6a shows the prediction results of ln(β) using a single BNN model and msBNN model, under the accumulation of errors, the msBNN model is of higher accuracy compared to BNN. It should be noted that we only used very little data, and under sufficient data conditions, such a theory-guided model will perform even better.

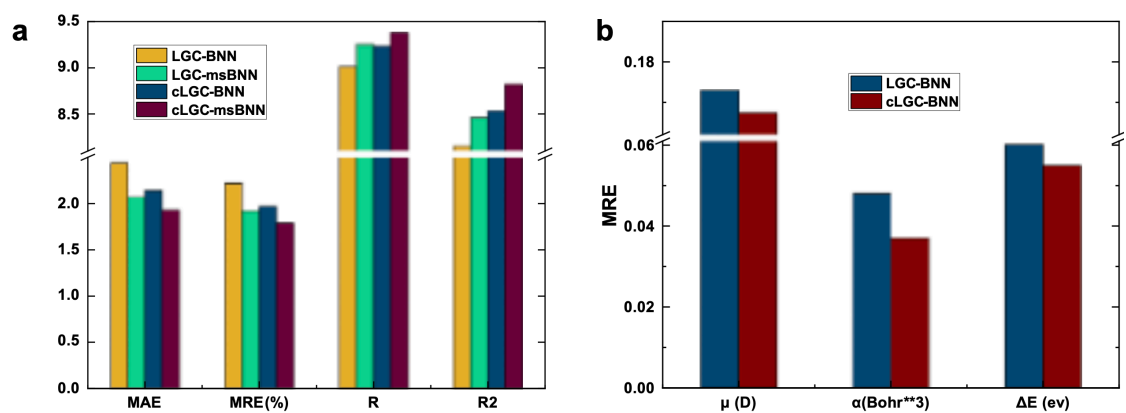

**Figure 6**. The prediction results of ln(β) using different models in the test set (a) Prediction results of different models for ln(β). (b) Prediction results of different models for *μ*, *α*, and *ΔE*.

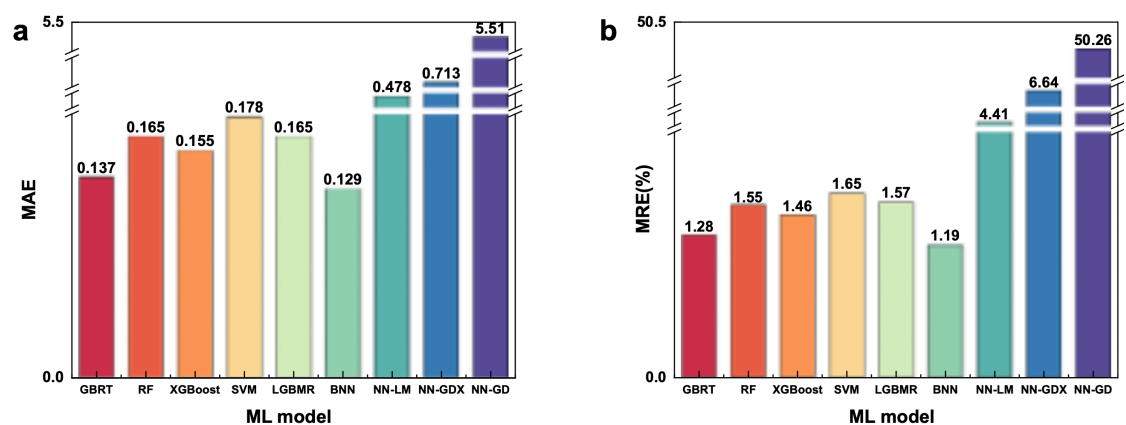

**Figure 7**. The prediction results of ln(β) using different models in the test set (all models use the LGC method as input). (a) Mean absolute error (MAE) of different models. (b) Mean relative error (MRE%) of different models.

For further comparison, we analyzed the results by different models (Including: Random forests (RF)[31], Gradient Boosting Regression Tree (GBRT)[32], eXtreme Gradient Boosting (XGBoost)[33], Light Gradient Boosting Machine Regression (LGBMR)[34], Support vector regression (SVR) and neural network models based on other algorithms (Levenberg-Marquardt backpropagation (LM)/Gradient descent with momentum and adaptive learning rate backpropagation (GDX)/Gradient descent backpropagation (GD)). As shown in Figure 7a and Figure 7b, the performance of Bayesian neural networks (BNN) is not only superior to other machine learning models, but also far superior to other neural networks. Most likely, all input features cannot contribute independently to the results, and Bayesian neural networks are better in line with mathematical rules; it is thus more capable of fitting correct chemical rules, and less prone to overfit compared to other neural networks.

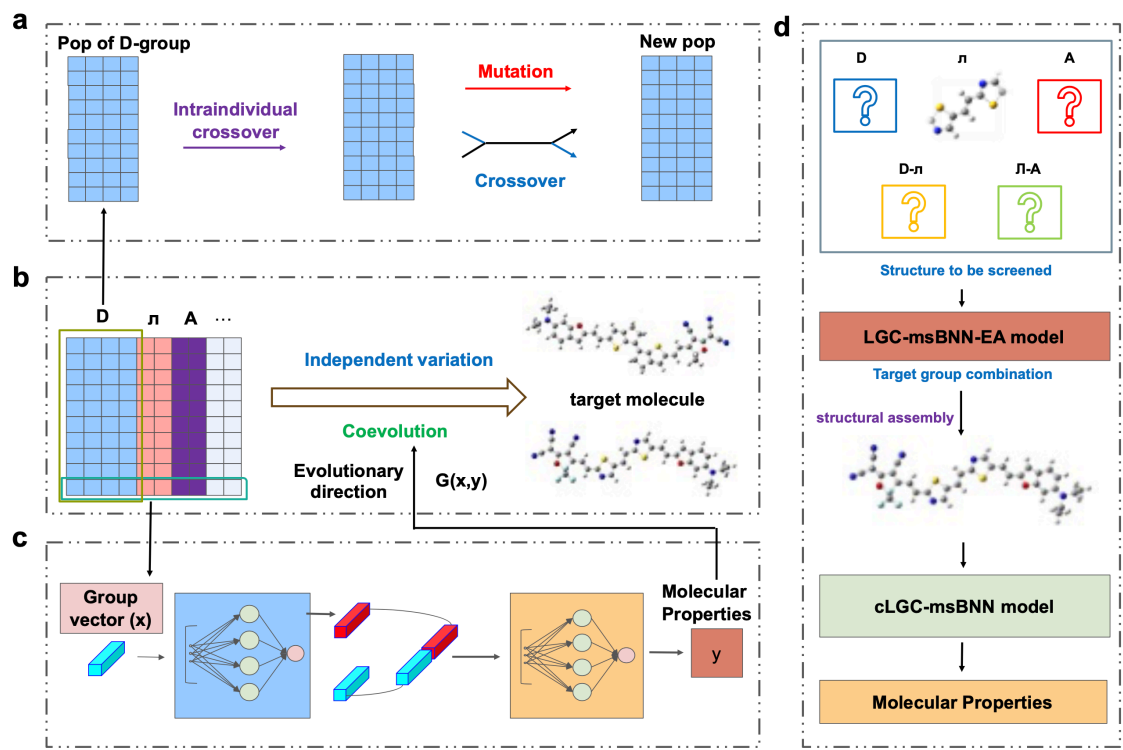

**Figure 8.** Structure of LGC-msBNN-EA model cLGC–msBNN model. (a) The process of cross mutation of electron donor groups during evolution. (b) The evolution process

of group population. (c) The coordination process between evolutionary algorithm and LGC-msBNN model. (d) Assembly of structures generated by LGC-msBNN-EA model and prediction through cLGC–msBNN model.

**Model application**

After establishing the prediction model, we combined the LGC-msBNN model with evolutionary algorithms to design target molecules, and the detailed process is shown in Figure 8. The initial population matrix contains hundreds of group vectors ($\vec{Ng}$). Upon crossover and mutation, the ln(β) values of evolved individuals are calculated using the msBNN model, and subsequently sorted out by the fitness functions $G(x,y)$. The latter consists of molecular structure function ($f(\vec{x})$) and molecular property function ($f(y)$). With continuous evolution, the ideal group vector will be obtained eventually. The fitness function is as follows:

$$f(\vec{x}) = f(\vec{n_\pi}, \vec{n_{D\text{-}\pi}}, \vec{n_{\pi\text{-}A}}, \vec{n_A}, \vec{n_D}) \tag{16}$$

$$f(y) = f(\ln(\beta)) \tag{17}$$

$$G(x,y) = \frac{1}{1/f(y) + f(\vec{x})} \tag{18}$$

in which $f(y)$ is a function making judgement according to the target optical properties, while $f(\vec{x})$ is used to determine whether the structure is the needed one. If the evolved structure does not meet initial setting, the fitness function $G(x,y)$ approaches 0. In order to make the evolutionary algorithm more suitable for input features, different group regions are crossed separately (Figure 6a). This not only greatly increases the interpretability of the model, but also improves the search efficiency, making it easier to jump out of the local optimal solution. As shown in Figure 9, we compared the search efficiency of our method with the one with global crossover manner; apparently, the latter is prone to be trapped in local optima.

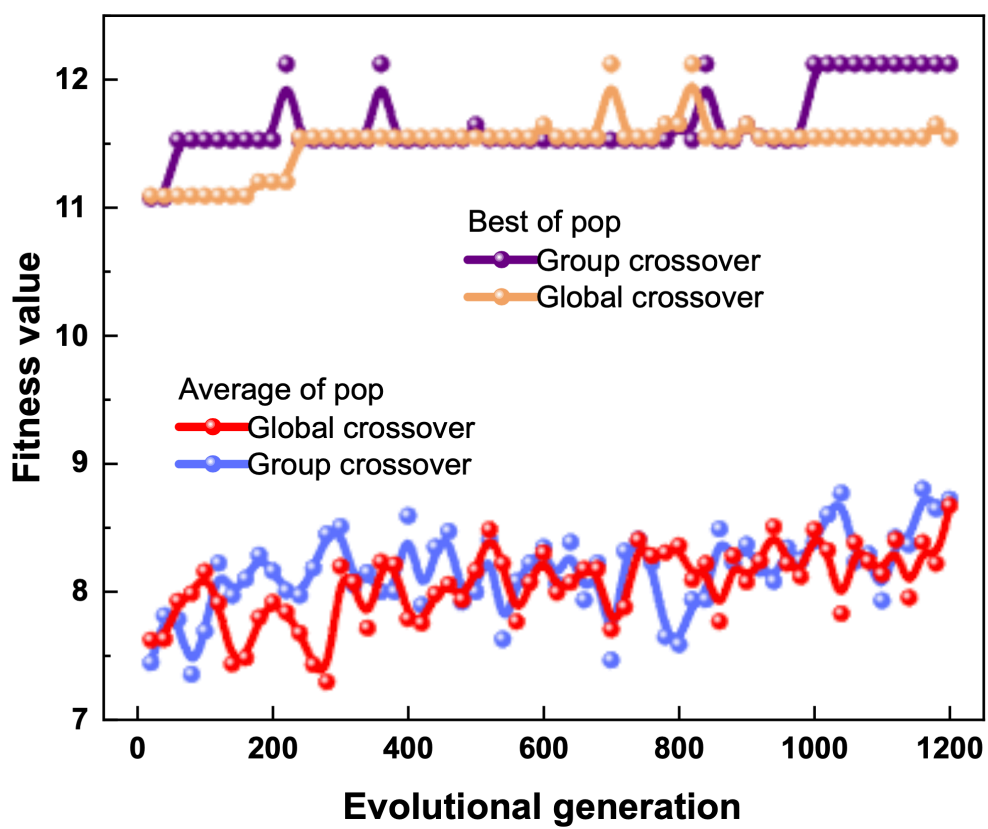

**Figure 9**. Trends in fitness values of different crossover methods during an evolution.

Such a target-driven molecular design framework has strong flexibility, especially for the generation of certain type of molecules. As shown in Figure 4, when the electron donors/ acceptors is connected to the bridge groups with a >C=C<, the molecule tends to have a higher first-order hyperpolarizability. In order to increase search efficiency, during the course of molecular design, the connecting group was fixed as >C=C< to generate molecules. Based on this, three molecules were discovered with high first-order polarizability, and their structural formulas and properties are shown in Table 3 and Table 4. To verify the accuracy of our prediction model in practical applications, we used DFT to calculate the properties of these three molecules. As shown in Table 3 and Table 4, although the model has gone through so many stages, the prediction results agree well with DFT calculation. Moreover, the first-order hyperpolarizabilities of the three molecules discovered are leading the dataset, indicating that the cLGC-msBNN model is capable of accurate prediction even the target feature is beyond the range of the dataset.

**Table 3.** The structural formulas of three discovered molecules.

| Number | molecular structure |
|---|---|
| 1 | 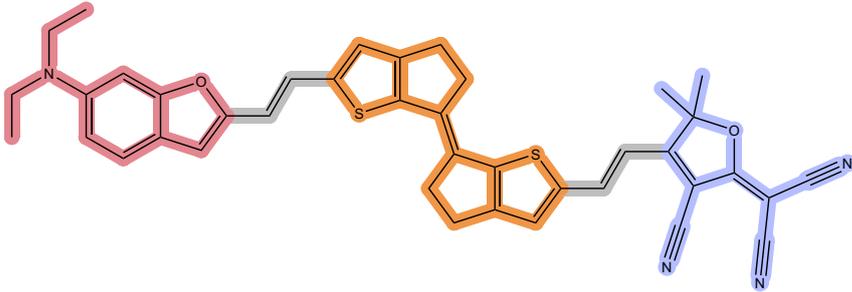 |
| 2 | 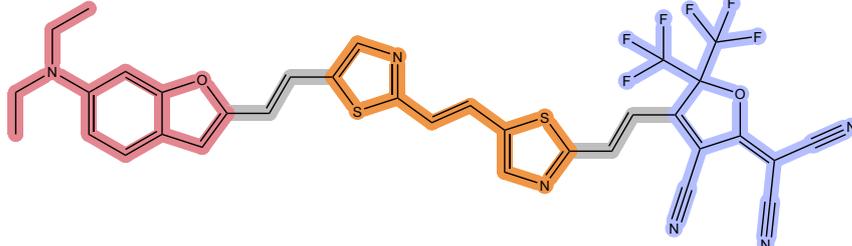 |
| 3 | 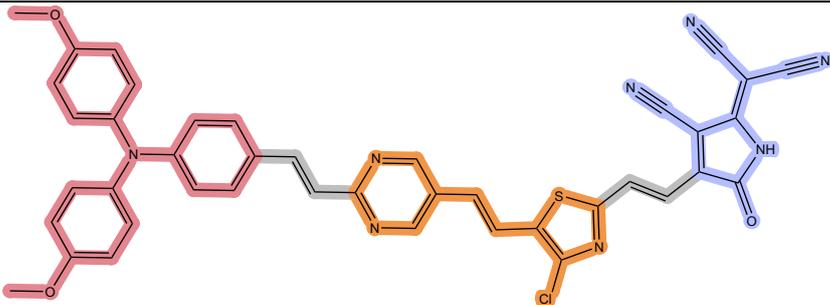 |

**Table 4.** Various properties of the three molecules discovered.

| Number | $\mu$ (D) (DFT/pre) | $\alpha$ (Bohr**3) (DFT/pre) | $\Delta E$(ev) (DFT/pre) | $\beta \cdot 10^{-30}$(esu) (DFT/pre) | $\ln(\beta)$ (DFT/pre) |
|---|---|---|---|---|---|
| 1 | 21.44 (22.44) | 1030 (996.2) | 1.613 (1.640) | 1799 (1420) | 12.25 (12.01) |
| 2 | 19.10 (17.18) | 942.5 (904.4) | 1.163 (1.120) | 2599 (2015) | 12.61 (12.36) |
| 3 | 12.99 (12.71) | 1006 (993.5) | 0.9865 (0.995) | 2387 (2205) | 12.53 (12.45) |

Undoubtedly, the DFT calculation have certain errors, while the combination of BNN and cLGC accurately repeat the DFT results with a small number of samples.

Thus, both the prediction of target feature and the generation of wanted molecule depend on the accuracy of the initial data. In more general backgrounds, with small sample of reliable data, this molecular design framework may fit broader physical/chemical space for small organic molecules.

## CONCLUSION

In summary, we develop an interactive molecular design framework to accelerate the discovery of NLO materials. The architecture of the framework is based on the LGC and cLGC methods designed for D-π-A type molecules. The msBNN model was employed for predicting various optical properties, affording high accuracy with a rather small data set for training. Further, by combining the prediction model and the evolutionary algorithm specially adapted for the LGC method, molecular design was achieved with minimal data. Such a method is expected to remedy the defect of machine learning model when data is deficient, and may thus prove generally applicable in molecular design for wider chemical/physical aims.

## DATA AND METHOD

**DFT calculation**
The DFT calculation was performed using the Gaussian 16[35] program. All structures were optimized at the B3LYP[36, 37]-D3(BJ)[38, 39]/def2-TZVP[40, 41] level of theory. The $\mu$ and $\alpha$ values were obtained at the cam-B3LYP[42]-D3(BJ)/def2-TZVP level of theory, while $\Delta E$ at B3LYP-D3(BJ)/def2-TZVP. Stationary points were optimized without symmetry constraint, and their nature was confirmed by vibrational frequency analysis. Unscaled vibrational frequencies were used to correct the relative energies for zero-point vibrational energy (ZPVE) contributions.

**GC method**

**Table 5.** List of parameters of the group contribution method.

| First-order feature |
|---|
| **D** |
| 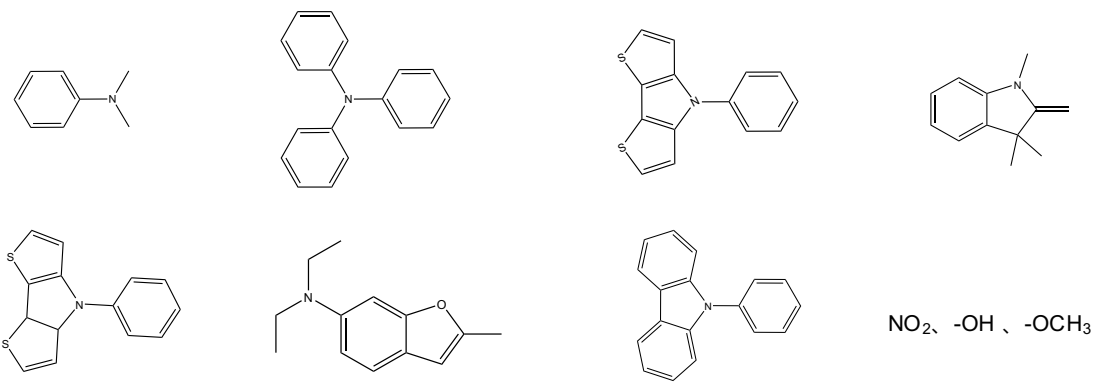 NO₂、-OH、-OCH₃ |
| **π** |
| 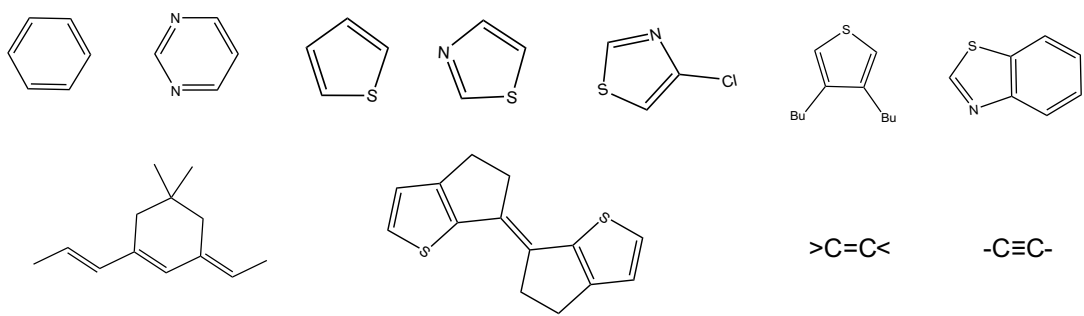 >C=C<    -C≡C- |
| **A** |

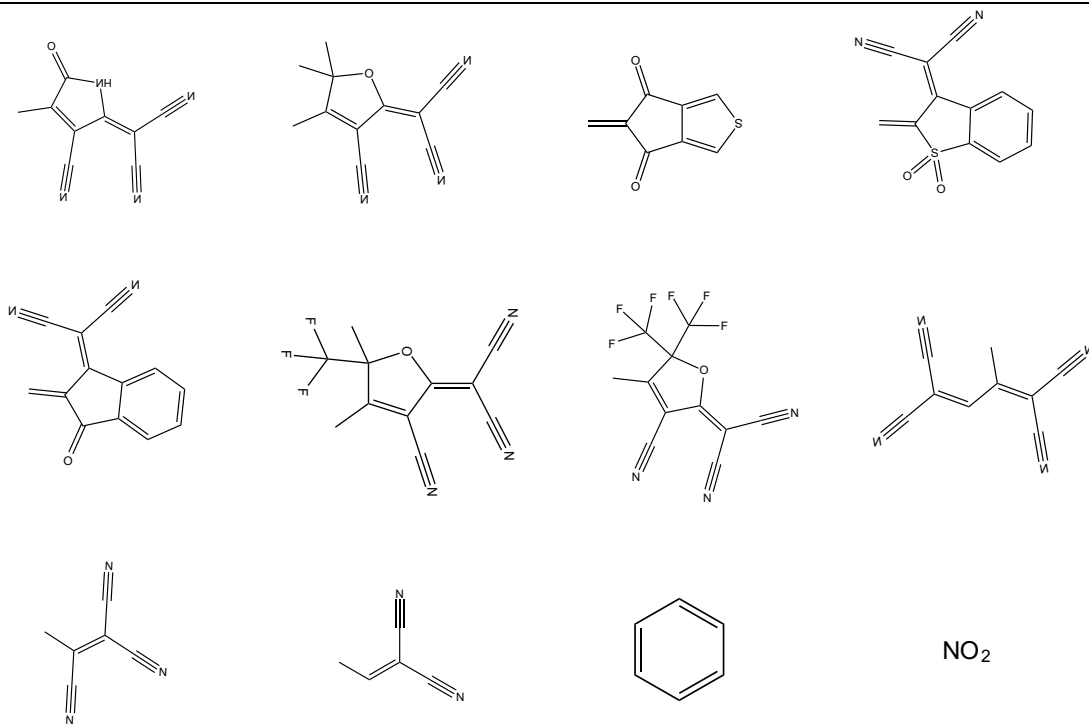

NO₂

| **D-π** | | |
|---|---|---|
| -C=C- | -Ring | -C≡C- |
| **π-A** | | |
| -C=C- | -Ring | -C≡C- |

| Second-order feature | | |
|---|---|---|
| D | π | A |
| C_D | C_π | C_A |
| >C=C< | N(Thiazole) | >C=C< |
| N(6ring) | N(benzene) | -CN3 |
| N(5ring) | N(Thiophene) | N(TCF) |
| -N<(-Ring) | | N(6ring) |
| -N<(Ring) | | N(5ring) |
| | | C=O |
| | | C=C |
| | | -C≡N |

With this method, the structural characteristics of molecules can be converted into computer readable form. Some descriptors are explained as follows:

>C=C<: indicates that the aromatic ring is connected with the aromatic ring.

N(6ring): Number of six component aromatic ring.

N(5ring): Number of six component aromatic ring.

-N<(-Ring): indicates -N< connected with aromatic ring.

-N<(Ring): Representing -N< participating in ring formation.

N(TCF): The number of tricyanofuran (TCF) with different substituents (such as $CF_3$-TCF).

C=O(Ring): Representing C=O participating in ring formation.

C=C: C=C with electron acceptors and electron bridges connected at both ends.

**Third-order features**:

(1) Electrotopological state (E-state) descriptors: This index combines the electronic states of intramolecular bonding atoms and their topological properties in the whole molecular skeleton. According to this descriptor, three internal states of the molecular substructure within the molecule are quantified: its element content, its valence state (electronic organization) and its topological state relative to its atomic neighbor.[43]

(2) Molecular operating environment (MOE-type) descriptors: The MOE-type descriptors use connectivity information and van der Waals radii to calculate the atomic van der Waals surface area (VSA) contribution of an atom-type to a given property. Including polarizability, direct electrostatic interaction and other factors.[44]

(3) Topological descriptors: According to this descriptor, the connection state of each

atom is used to calculate the exponent, thus providing a highly unique exponent for a given molecule.[45]

(4) Connectivity descriptors.

**Machine learning model.**

Bayesian neural network (BNN): Deep neural network often uses gradient descent algorithm with fast convergence in realizing nonlinear regression. For small sample learning, one can sacrifice the convergence speed to obtain high accuracy. On the other hand, due to the strong learning ability of neural network, over fitting is commonly encountered when there are few data. Therefore, we use Bayesian regularization to reduce over fitting and improve the generalization ability of the neural networks. Moreover, this neural network can achieve ideal results without the need for normalization, which will make it more adaptable to evolutionary algorithms. In the Bayesian framework the weights of the network are considered as random variables. After the data is taken, the density function for the weights can be updated according to Bayes' rule: [46,47]

$$P(w|D,\alpha,\beta,M) = \frac{P(D|w,\beta,M)P(w|\alpha,M)}{P(D|\alpha,\beta,M)} \quad (19)$$

in which $D$ represents the data set; $M$ is the particular neural network model used; $w$ is the vector of network weights; $P(w|\alpha,M)$ is the prior density, i.e. the known weights before any data is collected; $P(D|w,\beta,M)$ is the likelihood function, i.e. the probability of the data occurring under the weights $w$; $P(D|\alpha,\beta,M)$ is a normalization factor, which guarantees the total probability of 1.[47] The Hessian matrix of Gauss Newton approximation can be implemented within the framework of Levenberg Marquardt algorithm to reduce the computing cost.

We use the following indicators to evaluate the prediction performance of the model: mean relative error (MRE), mean absolute error (MAE), Pearson correlation coefficient (R), and the square of Pearson correlation coefficient (R2), the formula for calculating the error is as follows:

$$\text{Error} = y_{DFT} - y_{pred} \quad (20)$$

$$\text{MSE} = \frac{1}{n}\sum_{t=1}^{n}(y_{\text{Pred}} - y_{\text{DFT}})^2 \tag{21}$$

$$\text{MAE} = \frac{1}{n}\sum_{t=1}^{n}\left|y_{\text{DFT}} - y_{\text{pred}}\right| \tag{22}$$

$$\text{MRE} = \frac{1}{n}\sum_{t=1}^{n}\left|\frac{y_{\text{DFT}} - y_{\text{Pred}}}{y_{\text{DFT}}}\right| \tag{23}$$

In the formula, $y_{\text{DFT}}$ is the value calculated by DFT, and $y_{\text{pred}}$ is values predicted by neural networks.

**Evolutionary algorithm.**

In the initial population of evolutionary algorithm (we set it to 550 individuals, and the number can be adjusted according to needs), each individual is composed of binary vectors with a length of 28, the vector lengths of different groups are:1) G_D:7. 2) G_п:9. 3) G_A:11. The remaining groups are set as constants and participate in subsequent model predictions and fitness function calculations along with variables in the individual. By changing the fitness function, molecules containing different numbers of functional groups can be obtained. This can also set G_D/G_п/G_A in the initial population as a constant to find the optimal group combination.

The data and more details about the framework are available at https://github.com/Fan1ing/LGC-msBNN-EA.

# ACKNOWLEDGEMENT

Generous financial support by the *Zhejiang Provincial "Jianbing" and "Lingyan" R&D Programs* (2023C01102, 2023C01208).

# REFERENCES

1. Butler, K. T.; Davies, D. W.; Cartwright, H.; Isayev, O.; Walsh, A. Machine learning for molecular and materials science. *Nature.* **2018,** *559* (7715), 547-555.
2. Granda, J. M.; Donina, L.; Dragone, V.; Long, D. L.; Cronin, L. Controlling an organic synthesis robot with machine learning to search for new reactivity. *Nature.* **2018,** *559* (7714), 377-381.
3. Zhong, M.; Tran, K.; Min, Y.; Wang, C.; Wang, Z.; Dinh, C.-T.; De Luna, P.; Yu, Z.; Rasouli, A. S.; Brodersen, P.; Sun, S.; Voznyy, O.; Tan, C.-S.; Askerka, M.; Che, F.; Liu, M.; Seifitokaldani, A.; Pang, Y.; Lo, S.-C.; Ip, A.; Ulissi, Z.; Sargent, E. H.

Accelerated discovery of CO2 electrocatalysts using active machine learning. *Nature.* **2020,** *581* (7807), 178-183.
4. Gu, G. H.; Noh, J.; Kim, I.; Jung, Y. Machine learning for renewable energy materials. *J. Mater. Chem. A.* **2019,** *7* (29), 17096-17117.
5. Masood, H.; Toe, C. Y.; Teoh, W. Y.; Sethu, V.; Amal, R. Machine Learning for Accelerated Discovery of Solar Photocatalysts. *ACS Catal.* **2019,** *9* (12), 11774-11787.
6. Gebauer, N. W. A.; Gastegger, M.; Hessmann, S. S. P.; Muller, K. R.; Schutt, K. T. Inverse design of 3d molecular structures with conditional generative neural networks. *Nat. Commun.* **2022,** *13* (1), 973.
7. Guo, S.; Popp, J.; Bocklitz, T. Chemometric analysis in Raman spectroscopy from experimental design to machine learning-based modeling. *Nat. Protoc.* **2021,** *16* (12), 5426-5459.
8. Soleimany, A. P.; Amini, A.; Goldman, S.; Rus, D.; Bhatia, S. N.; Coley, C. W. Evidential Deep Learning for Guided Molecular Property Prediction and Discovery. *ACS Cent. Sci.* **2021,** *7* (8), 1356-1367.
9. Ulissi, Z. W.; Medford, A. J.; Bligaard, T.; Norskov, J. K. To address surface reaction network complexity using scaling relations machine learning and DFT calculations. *Nat. Commun.* **2017,** *8*, 14621.
10. Yang, Z.; Buehler, M. J. Linking atomic structural defects to mesoscale properties in crystalline solids using graph neural networks. *npj Comput. Mater.* **2022,** *8* (1).
11. Bødker, M. L.; Bauchy, M.; Du, T.; Mauro, J. C.; Smedskjaer, M. M. Predicting glass structure by physics-informed machine learning. *npj Comput. Mater.* **2022,** *8* (1).
12. Yang, G.; Wu, K. Two-dimensional nonlinear optical materials predicted by network visualization. *Mol. Syst. Des. Eng.* **2019,** *4* (3), 586-596.
13. Lemm, D.; von Rudorff, G. F.; von Lilienfeld, O. A. Machine learning based energy-free structure predictions of molecules, transition states, and solids. *Nat. Commun.* **2021,** *12* (1), 4468.
14. Gomez-Bombarelli, R.; Wei, J. N.; Duvenaud, D.; Hernandez-Lobato, J. M.; Sanchez-Lengeling, B.; Sheberla, D.; Aguilera-Iparraguirre, J.; Hirzel, T. D.; Adams, R. P.; Aspuru-Guzik, A. Automatic Chemical Design Using a Data-Driven Continuous Representation of Molecules. *ACS Cent. Sci.* **2018,** *4* (2), 268-276.
15. Popova, M.; Isayev, O.; Tropsha, A. Deep reinforcement learning for de novo drug design. *Sci. Adv.* **2018,** *4* (7), eaap7885.
16. Cao, Y.; Romero, J.; Olson, J. P.; Degroote, M.; Johnson, P. D.; Kieferova, M.; Kivlichan, I. D.; Menke, T.; Peropadre, B.; Sawaya, N. P. D.; Sim, S.; Veis, L.; Aspuru-Guzik, A. Quantum Chemistry in the Age of Quantum Computing. *Chem. Rev.* **2019,** *119* (19), 10856-10915.
17. Fecková, M.; le Poul, P.; Bureš, F.; Robin-le Guen, F.; Achelle, S. Nonlinear optical properties of pyrimidine chromophores. *Dyes Pigm.* **2020,** *182*.
18. Semin, S.; Li, X.; Duan, Y.; Rasing, T. Nonlinear Optical Properties and Applications of Fluorenone Molecular Materials. *ADV OPT MATER.* **2021,** *9* (23).
19. Klikar, M.; le Poul, P.; Ruzicka, A.; Pytela, O.; Barsella, A.; Dorkenoo, K. D.; Robin-le Guen, F.; Bures, F.; Achelle, S. Dipolar NLO Chromophores Bearing Diazine Rings as pi-Conjugated Linkers. *J Org Chem.* **2017,** *82* (18), 9435-9451.

20. Enekvist, M.; Liang, X.; Zhang, X.; Dam-Johansen, K.; Kontogeorgis, G. M. Estimating Hansen solubility parameters of organic pigments by group contribution methods. *Chin. J. Chem. Eng*. **2021,** *31*, 186-197.

21. Cibulka, I.; Hnědkovský, L. Group contribution method for standard molar volumes of aqueous aliphatic alcohols, ethers and ketones over extended ranges of temperature and pressure. *J. Chem. Thermodyn.* **2011,** *43* (8), 1215-1223.

22. Fonseca, L. A. A. P.; Cremasco, M. A. Group contribution methods to predict enthalpy of vaporization of aromatic and terpene ketones at 298.15 K. *Fluid Phase Equilibria.* **2021,** *538*

23. LeCun, Y.; Bengio, Y.; Hinton, G. Deep learning. *Nature.* **2015,** *521* (7553), 436-44.

24. Fan, J.; Qian, C.; Zhou, S. Machine Learning Spectroscopy Using a 2-Stage, Generalized Constituent Contribution Protocol. *Research.* **2023**, *6*, 0115.

25. Kudela, J. A critical problem in benchmarking and analysis of evolutionary computation methods. *Nat. Mach. Intell.* **2022,** *4* (12), 1238-1245.

26. Li, B.; Wei, Z.; Wu, J.; Yu, S.; Zhang, T.; Zhu, C.; Zheng, D.; Guo, W.; Zhao, C.; Zhang, J. Machine learning-enabled globally guaranteed evolutionary computation. *Nat. Mach. Intell.* **2023,** *5* (4), 457-467.

26. Breitung, E. M.; Shu, C.-F.; McMahon, R. J. Thiazole and Thiophene Analogues of Donor−Acceptor Stilbenes:  Molecular Hyperpolarizabilities and Structure−Property Relationships. *J. Am. Chem. Soc.* **2000,** *122* (6), 1154-1160.

27. Marder, S. R.; Cheng, L.-T.; Tiemann, B. G.; Friedli, A. C.; Blanchard-Desce, M.; Perry, J. W.; Skindhøj, J. Large First Hyperpolarizabilities in Push-Pull Polyenes by Tuning of the Bond Length Alternation and Aromaticity. *Science.* **1994,** *263* (5146), 511-514.

28. Lu, D.; Chen, G.; Perry, J. W.; Goddard, W. A., III. Valence-Bond Charge-Transfer Model for Nonlinear Optical Properties of Charge-Transfer Organic Molecules. *J. Am. Chem. Soc.* **1994,** *116* (23), 10679-10685.

29. Meyers, F.; Marder, S. R.; Pierce, B. M.; Bredas, J. L. Electric Field Modulated Nonlinear Optical Properties of Donor-Acceptor Polyenes: Sum-Over-States Investigation of the Relationship between Molecular Polarizabilities (.alpha., .beta., and .gamma.) and Bond Length Alternation. *J. Am. Chem. Soc.* **1994,** *116* (23), 10703-10714.

30. Matlock, M. K.; Dang, N. L.; Swamidass, S. J. Learning a Local-Variable Model of Aromatic and Conjugated Systems. *ACS Cent. Sci.* **2018,** *4* (1), 52-62.

31. Breiman, L. Random forests. *Mach. Learn.* **2001,** *45*, 5−32.

32. Friedman, J. H. Greedy function approximation: a gradient boosting machine. *Ann. Stat.* **2001,** *29*, 1189−1232.

33. Chen, T.; Guestrin, C. XGBoost: a Scalable Tree Boosting System. *arXiv*:*1603.02754v3,* **2016.**

34. Ke, G.; Meng, Q.; Finley, T.; Wang, T.; Chen, W.; Ma, W.;Ye, Q.;Liu, T. LightGBM: a highly efficient gradient boosting decision tree. NIPS'17: *Proceedings of the 31st International Conference on Neural Information Processing Systems*; Curran Associates Inc.: California, USA, New York, Dec 4–9, **2017**; pp3149–3157.

35. Gaussian 16, Revision C.01, Frisch, M. J.; Trucks, G. W.; Schlegel, H. B.; Scuseria, G. E.; Robb, M. A.; Cheeseman, J. R.; Scalmani, G.; Barone, V.; Petersson, G. A.; Nakatsuji, H.; Li, X.; Caricato, M.; Marenich, A. V.; Bloino, J.; Janesko, B. G.; Gomperts, R.; Mennucci, B.; Hratchian, H. P.; Ortiz, J. V.; Izmaylov, A. F.; Sonnenberg, J. L.; Williams-Young, D.; Ding, F.; Lipparini, F.; Egidi, F.; Goings, J.; Peng, B.; Petrone, A.; Henderson, T.; Ranasinghe, D.; Zakrzewski, V. G.; Gao, J.; Rega, N.; Zheng, G.; Liang, W.; Hada, M.; Ehara, M.; Toyota, K.; Fukuda, R.; Hasegawa, J.; Ishida, M.; Nakajima, T.; Honda, Y.; Kitao, O.; Nakai, H.; Vreven, T.; Throssell, K.; Montgomery, J. A., Jr.; Peralta, J. E.; Ogliaro, F.; Bearpark, M. J.; Heyd, J. J.; Brothers, E. N.; Kudin, K. N.; Staroverov, V. N.; Keith, T. A.; Kobayashi, R.; Normand, J.; Raghavachari, K.; Rendell, A. P.; Burant, J. C.; Iyengar, S. S.; Tomasi, J.; Cossi, M.; Millam, J. M.; Klene, M.; Adamo, C.; Cammi, R.; Ochterski, J. W.; Martin, R. L.; Morokuma, K.; Farkas, O.; Foresman, J. B.; Fox, D. J. Gaussian, Inc., Wallingford CT, 2016.

36. Becke, A. D., A new mixing of Hartree–Fock and local density-functional theories. *J. Chem. Phys.* **1993,** *98* (2), 1372-1377.

37. Stephens, P. J.; Devlin, F. J.; Chabalowski, C. F.; Frisch, M. J. Ab Initio Calculation of Vibrational Absorption and Circular Dichroism Spectra Using Density Functional Force Fields. *J. Phys. Chem.* **1994,** *98* (45), 11623-11627.

38. Grimme, S.; Ehrlich, S.; Goerigk, L. Effect of the damping function in dispersion corrected density functional theory. *J. Comput. Chem.* **2011,** *32* (7), 1456-1465.

39. Grimme, S.; Antony, J.; Ehrlich, S.; Krieg, H. A consistent and accurate ab initio parametrization of density functional dispersion correction (DFT-D) for the 94 elements H-Pu. *J. Chem. Phys.* **2010,** *132* (15), 154104.

40. Weigend, F.; Ahlrichs, R. Balanced basis sets of split valence, triple zeta valence and quadruple zeta valence quality for H to Rn: Design and assessment of accuracy. *Phys. Chem. Chem. Phys.* **2005,** *7* (18), 3297-3305.

41. Andrae, D.; Häußermann, U.; Dolg, M.; Stoll, H.; Preuß, H. Energy-adjustedab initio pseudopotentials for the second and third row transition elements. *Theor. Chim. Acta.* **1990,** *77* (2), 123-141.

42. T. Yanai, D. Tew, and N. Handy, "A new hybrid exchange-correlation functional using the Coulomb-attenuating method (CAM-B3LYP)," *Chem. Phys. Lett.* **393** (2004) 51-57.

43. L, H, Hall.; L, B, Kier. Electrotopological state indices for atom types: A novel combination of electronic, topological, and valence state information. *J. Chem. Inf. Model*. **1995,** *35* (6), 1039–104.

44. P, A, Labute. widely applicable set of descriptors. J. Mol. Graph. Model. **2000,** *18* (4–5), 464–477.

45. H, Narumi. New topological indices for finite and infinite systems, Commun Math. *J. Comput. Chem.* **1987,** *22*, 195–207.

46. MacKay.; David, J, C. Bayesian interpolation. *Neural Comput.* **1992,** *4* (5), 415–447.

47. F, Dan, Foresee.; M, T, Hagan. Gauss-Newton approximation to Bayesian learning. *Proc. Int. Jt. Conf. Neural Netw*. **1997,** *3*, 1930-1935.